\documentclass[10pt,twocolumn,letterpaper]{article}

\usepackage{cvpr}              % To produce the CAMERA-READY version
\usepackage{graphicx}
\usepackage{balance}
\usepackage{booktabs} % For formal tables
\usepackage{placeins}
\usepackage[dvipsnames]{xcolor}

\usepackage[pagebackref,breaklinks,colorlinks]{hyperref}
\usepackage{float}
\newcommand{\qheading}[1]{\noindent\mbox{\textbf{#1}}}
\newcommand{\RNum}[1]{\uppercase\expandafter{\romannumeral #1\relax}}
\definecolor{DeltaColor}{rgb}{0.039,0.73,0.71}
\definecolor{SigmaColor}{rgb}{0.98,0.45,0.0}
\definecolor{AlphaColor}{rgb}{0,0,0.8}
\definecolor{BetaColor}{rgb}{0.8,0,0.8}
\definecolor{GammaColor}{rgb}{0.514,0.34,0.224}
\definecolor{EpsilonColor}{rgb}{0.353,0.725,0.906}
\definecolor{GreenColor}{rgb}{0.137,0.573,0.565}
\definecolor{RedColor}{rgb}{0.949,0.275, 0.224}
\definecolor{BlueColor}{rgb}{0.0,0.0, 0.99}
\definecolor{citecolor}{HTML}{0071bc}

\title{OutfitAnyone: Ultra-high Quality Virtual Try-On for Any Clothing and Any Person}

\author{%
\textbf{Ke Sun}$^{1*}$ \quad \textbf{Jian Cao}$^{1*}$ \quad \textbf{Qi Wang}$^{1}$ \quad \textbf{Linrui Tian}$^1$ \quad \textbf{Xindi Zhang}$^1$ \quad
\textbf{Lian Zhuo}$^1$ \quad \\ \textbf{Bang Zhang}$^1$ \quad \textbf{Liefeng Bo}$^1$ \quad \textbf{Wenbo Zhou}$^{2}$ \quad \textbf{Weiming Zhang}$^{2}$ \quad \textbf{Daiheng Gao}$^{2,3}$\\
$^1$Intelligent Computing, Tongyi, Alibaba Group \quad $^2$USTC \quad $^3$Formation.ai
}

\begin{document}

\twocolumn[{%
\renewcommand\twocolumn[1][]{#1}%
\maketitle
\begin{center}
    \centering
    \captionsetup{type=figure}
    \includegraphics[width=1.0\textwidth]{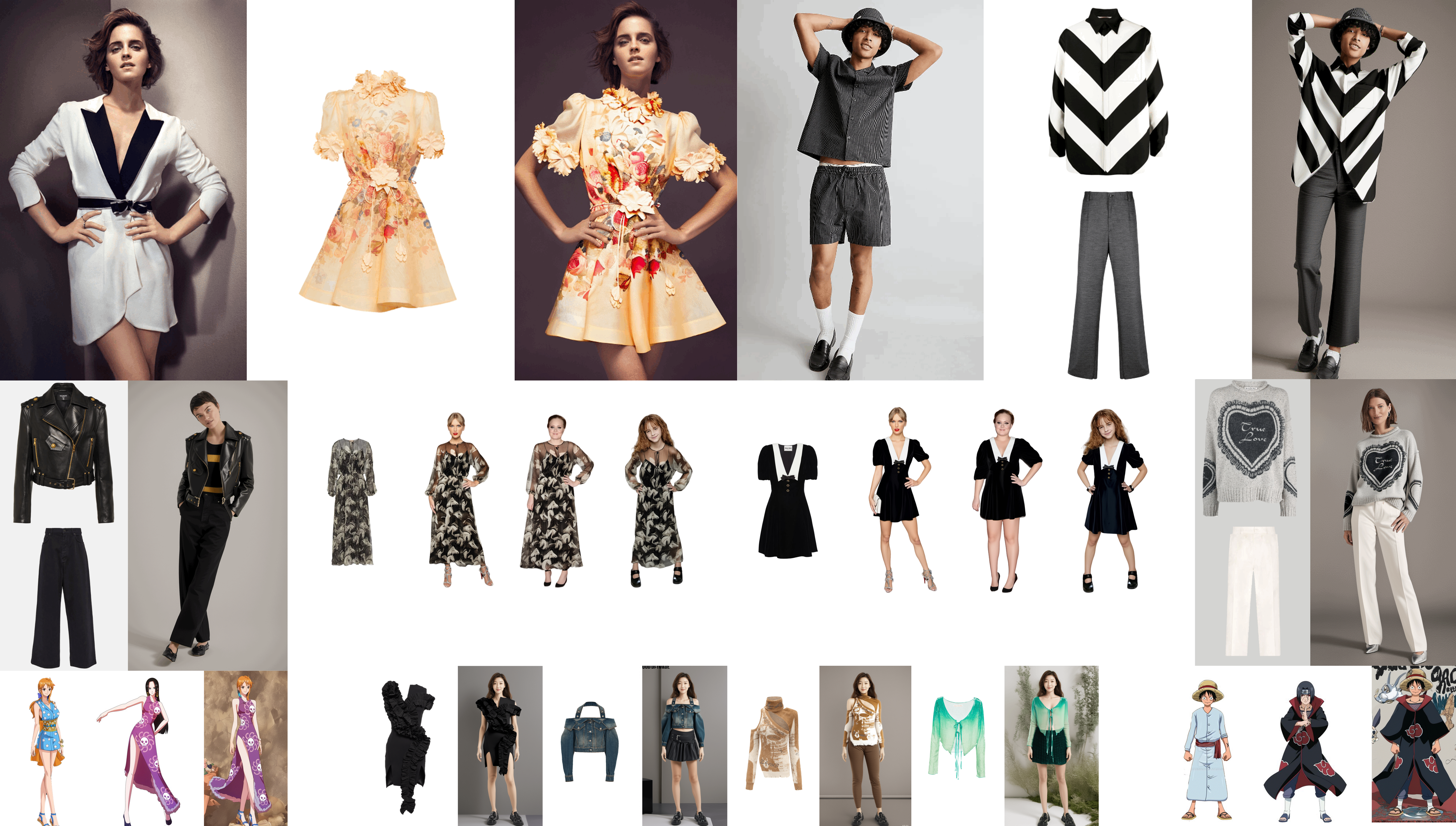}
    \captionof{figure}{We introduce \textbf{OutfitAnyone}, a diffusion-based framework for 2D Virtual Try-On. By far, it has garnered over 5,000 stars on GitHub and ranked within the top 20 among all the Hugging Face spaces.}
\end{center}%
}]

% \maketitle
% \begin{figure*}
% \includegraphics[width=1.0\linewidth]{fig/intro/teaser.png}
%   \caption{We propose \textbf{Cloth2Tex}, a novel pipeline for converting 2D images of clothing to high-quality 3D textured meshes that can be draped onto 3D humans. In contrast to previous methods, Cloth2Tex supports a variety of clothing types. Results of 3D textured meshes produced by our method as well as the corresponding input images are shown above.}
% \label{fig:teaser}
% \end{figure*}

\newcommand{\method}{OutfitAnyone\xspace}

\def\thefootnote{*}\footnotetext{These authors contributed equally to this work}

\begin{abstract}
Virtual Try-On (\textbf{VTON}) has become a transformative technology, empowering users to experiment with fashion without ever having to physically try on clothing. However, existing methods often struggle with generating high-fidelity and detail-consistent results. While diffusion models, such as Stable Diffusion 1/2/3, have shown their capability in creating high-quality and photorealistic images, they encounter formidable challenges in conditional generation scenarios like VTON. Specifically, these models struggle to maintain a balance between control and consistency when generating images for virtual clothing trials.

\textbf{OutfitAnyone} addresses these limitations by leveraging a two-stream conditional diffusion model, enabling it to adeptly handle garment deformation for more lifelike results. It distinguishes itself with scalability—modulating factors such as pose, body shape and broad applicability, extending from anime to in-the-wild images. OutfitAnyone's performance in diverse scenarios underscores its utility and readiness for real-world deployment. For more details and animated results, please see \url{https://humanaigc.github.io/outfit-anyone/}.
% Look at previous \confName abstracts to get a feel for style and length.
\end{abstract}
\section{Introduction}
\label{sec: intro}

The concept of Virtual Try-On (VTON) is centered around the ability to digitally simulate how a piece of clothing would appear on an individual, using their photograph and an image of the clothing item. This technology holds the promise of significantly enriching the online shopping experience. However, the effectiveness of most VTON methods is often limited to scenarios where there is minimal variation in body posture and shape. A critical unresolved challenge lies in the accurate non-rigid transformation of clothing to conform to a specific body shape without causing any distortion to the garment's patterns and textures ~\cite{choi2021viton, han2018viton, wang2018toward, gao2021shape}.

Currently, two primary approaches are being explored to address these challenges. The first is the template-based 3D Virtual Try-On (VTON), which has proven to be effective in tackling these issues, as demonstrated in various studies \cite{xu20193d, pix2surf, majithia2022robust, gao2024cloth2tex}. The underlying technology of these methods involves converting 2D images into 3D textures for clothing mesh models. The crux of creating 3D textures from 2D images lies in establishing accurate correspondences between the catalog images and the UV textures, either manually using techniques like the Thin-Plate-Spline (TPS) warping \cite{berg2006shape} or automatically through the As-Rigid-As-Possible (ARAP) deformation \cite{sorkine2007rigid} loss and the differentiable neural rendering~\cite{softras}.

While 3D Virtual Try-On (VTON) struggles to achieve the level of realism and diversity needed for a wide range of garments, it also faces the issue of longer processing times during inference, which is a significant disadvantage compared to 2D counterparts.

TryOnDiffusion~\cite{zhu2023tryondiffusion}, pioneering as the first VTON technique to harness the power of diffusion models, stands out as a quintessential algorithm in the second approach. It skillfully navigates significant occlusions, diverse poses, and alterations in body contours, meticulously maintaining the fine details of garments with high-resolution clarity. However, TryOnDiffusion can only solve single garment Try-On, which makes it utterly impractical for real-world usage.

Drawing inspiration from the innovative Parallel-UNet design featured in TryOnDiffusion, we've crafted OutfitAnyone, a cutting-edge technology dedicated to delivering ultra-high definition virtual try-ons for a wide array of clothing on any individual. It's the ultimate solution for handling large occlusions, a variety of poses and body shapes, and an extensive range of garments.

At the heart of our approach lies a conditional Diffusion Model that meticulously processes images of the model, the clothing, and accompanying textual prompts, harnessing garment images as a guiding influence. The architecture of the network is bifurcated into two distinct pathways, each independently handling the model and clothing data. These pathways merge within a sophisticated fusion network that adeptly integrates the intricacies of the garment onto the model.

 which encompasses two pivotal components: the Zero-shot Try-on Network that generates the initial try-on visuals, and the Post-hoc Refiner that meticulously refines the clothing and skin textures in the final output images.

In summary, we contribute \textbf{OutfitAnyone}, which encompasses two pivotal components: the Zero-shot Try-on Network that generates the initial try-on visuals, and the Post-hoc Refiner that meticulously refines the clothing and skin textures in the final output images. 

% Specifically, OutfitAnyone can synthesizing photorealistic Virtual Try-On on:
% \begin{itemize}
%     \item \textbf{Any Person}: please see \cref{fig:diff_outfits}, \cref{fig:kid} and \cref{fig:black_man}.
%     \item \textbf{Any Body Shape}: please see \cref{fig:shape} and our teaser.
%     \item \textbf{Any Style}: please see \cref{fig:bizzare}.
% \end{itemize}

Specifically, The characteristics of OutfitAnyone can be summarized as follows:
\begin{itemize}
    \item \textit{a)} \textbf{Cutting-edge Realism}: Our OutfitAnyone method sets a new industry standard for Virtual Try-On, delivering industry-leading, high-quality results.
    \item \textit{b)} \textbf{High Robustness}: The method we propose can support virtual try-on for anyone, any outfits, any body shape and any scenario.
    \item \textit{c)} \textbf{Flexible Control}:  We support various pose and body shape guidance methods, including (openpose~\cite{openpose}, SMPL~\cite{smpl}, densepose~\cite{densepose}).
    \item \textit{d)} \textbf{High Quality}: We support flexible sizes VTON synthesizing, from 384 (width) $\times$ 684 (height) to 1080 (width)  $\times$  1920 (height).
\end{itemize}

\section{Related Works}

\qheading{GAN-Based Virtual Try-on.} 
Traditional approaches to virtual try-on technology~\cite{wang2018toward,bai2022single,dong2019towards,he2022style,lee2022high,feng2022weakly} typically involve a two-step process that leverages Generative Adversarial Networks (GANs)~\cite{goodfellow2014generative}. Initially, a precise warping module is utilized to adjust the shape of in-shop clothing to match the desired silhouette. Subsequently, the second phase deploys a GAN-driven generator to seamlessly integrate the reshaped attire onto the subject's image.

The realism of these GAN-based methods hinges largely on the proficiency of the initial warping phase, prompting a concerted effort to bolster the warping module's proficiency in managing non-rigid deformations. Beyond this, the quality of GANs has faced some skepticism, particularly in the last two years, as diffusion models have showcased their prowessful capability to generate data from all kinds (image, video, 3D and etc).

\qheading{Diffusion-Based Virtual Try-On.} In contrast to GAN-based models, diffusion models have taken considerable leaps forward in the realm of high-fidelity conditional image generation~\cite{stablediffusion,podell2023sdxl,saharia2022photorealistic,controlnet,omost}. The process of image-based virtual try-on can be viewed as a specialized subset of the general image editing or restoration task, tailored to the specific conditions set by the provided garment and model.

DCI-VTON~\cite{gou2023taming} and LADI-VTON~\cite{morelli2023ladi} stand out as two seminal works that aim to bridge the gap between traditional GAN-based methodologies and diffusion models. These works utilize explicit warping modules to generate deformed garments and subsequently leverage diffusion models to seamlessly integrate these garments with images of the reference person.

TryonDiffusion~\cite{zhu2023tryondiffusion} has taken a bold step by forgoing the integration of garment warping modules into its process. This decision marks a significant departure from the traditional approach, as it removes the necessity for explicit warping and feature alignment mechanisms. It's commendable that TryonDiffusion has paved the way for an alternative in VTON technology, one that doesn't rely on the inclusion of meticulously crafted warping modules. This innovation has alleviated the substantial workload for researchers and engineers who previously had to invest significant effort in fine-tuning warping modules within the network—a task that has now become obsolete.

While TryonDiffusion has not yet fully exploited the potential of current Large Language Models (LLMs) and has opted to disregard text as an input method, MMTryOn~\cite{zhang2024mmtryon} adeptly leverages the prowess of expansive multi-modal models. This sophisticated strategy facilitates a more refined and versatile interaction with diverse apparel styles and Virtual Try-On contexts. Simultaneously, the concurrent research by OOTDiffusion~\cite{xu2024ootdiffusion} ingeniously employs CLIP~\cite{clip} to encode garment labels for upper body, lower body, and full-body try-on applications, enhancing the precision and adaptability of the system.
\begin{figure*}[t]
\includegraphics[width=1\linewidth]{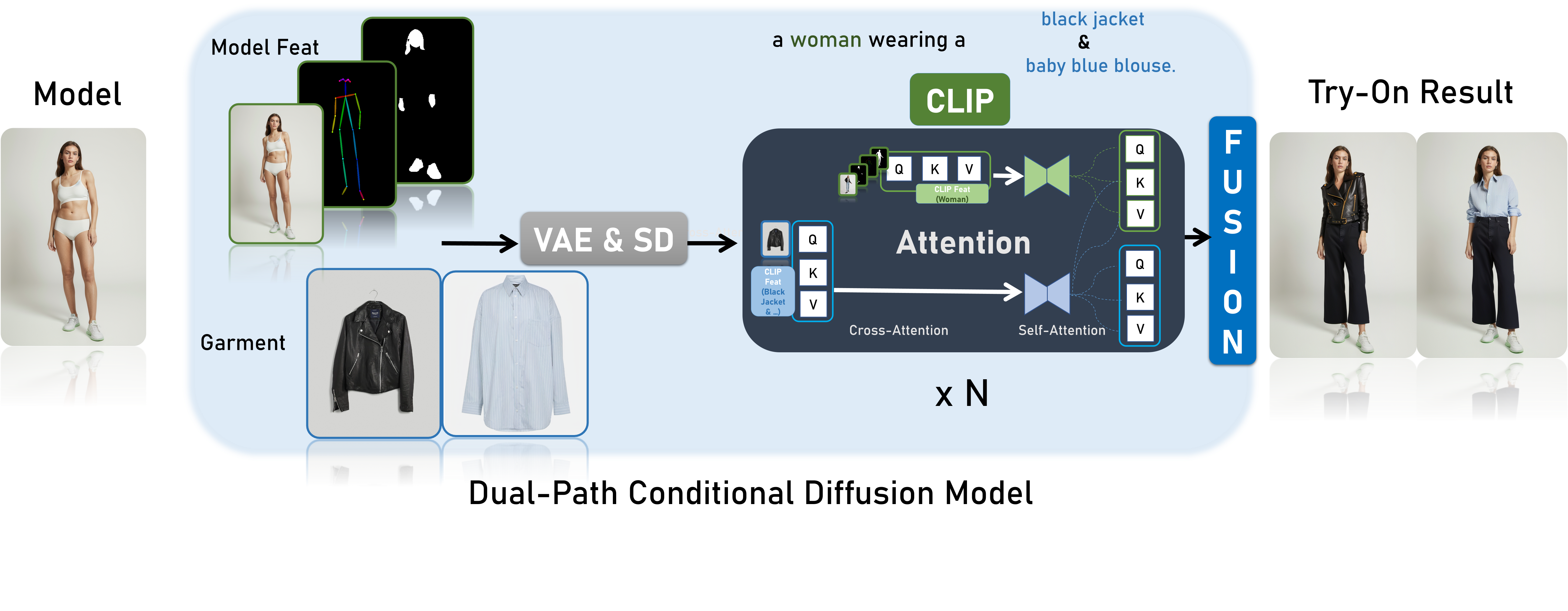}
  \caption{\textbf{Method overview}: \method processes input consisting of a model, garment, and related prompts through a dual-path conditional diffusion model. This model bifurcates into two distinct pathways, each dedicated to handling the model and garment data independently. The two streams eventually merge within a fusion network, which effectively integrates the garment details into the model's feature representation. To elaborate, we extract features: openpose (can be replaced by densepose or SMPL) and initmask from the model, and then concatenate these features with the model image. This composite data is then fed into our Dual-Path SD model, which guarantees not only the high-quality retention but also the restoration of the garment's features. Importantly, the feature spaces for both models and garments are aligned, which significantly accelerates the convergence process (with a visible try-on effect achievable within just 6k iterations). Significantly, the prompt, although it may not align perfectly with the spatial pixels, plays a crucial role in preserving semantic-level information.
}
  % In Phase \RNum{1}, we determine the 3D garment shape and coarse texture by registering our parametric garment meshes onto catalog images using a neural mesh renderer. Next, in Phase \RNum{2}, we refine the coarse estimate of the texture to obtain high-quality fine textures using image translation networks trained on large-scale data synthesized by pre-trained latent diffusion models. Note that the only component that requires training is the inpainting network. Please watch our project page for an animated explanation of \method.
\label{fig:maingraph}
\end{figure*}

\section{Overall Framework}
As shown in the \cref{fig:maingraph}, we developed a framework incorporating ReferenceNet, which effectively maintains the integrity of pattern and texture information from clothing images when they are used as conditions in the main generation pipeline. This design ensures that both the fit and visual details of the clothing are accurately preserved throughout the generation process. Internally, the network segregates into two streams for independent processing of model and garment data. These streams converge within a fusion network that facilitates the embedding of garment details onto the model’s feature representation. 
On this foundation, we have established OutfitAnyone, comprising two key elements: the Zero-shot Try-on Network for initial try-on imagery, and the Post-hoc Refiner for detailed enhancement of clothing and skin texture in the output images.

% The characteristics of OutfitAnyone are 4-folds:
% \begin{itemize}
%     \item \textit{a)} \textbf{Dataset}: Trained on 1 Million image pairs (model, garment, pose, mask).
%     \item \textit{b)} \textbf{Inference}: Can be deployed on NVIDIA A10 24G VRAM easily, average inference time is 5.7s.
%     \item \textit{c)} \textbf{Guidance}: Support all kinds of guidance (openpose~\cite{openpose}, smpl~\cite{smpl}, densepose~\cite{densepose}) for pose and bodyshape guidance.
%     \item \textit{d)} \textbf{Resolution}: Support flexible sizes VTON synthesizing, from 384 (width) $\times$ 684 (height) to 1080 (width)  $\times$  1920 (height).
% \end{itemize}
% \subsection{Stable Diffusion}
% \label{sec:phase1}
% % \xuchen{maybe Silhouette alignment is more precise}

\subsection{Clothing Feature Injection}
\label{sec:phase30}
Stable Diffusion (SD)~\cite{stablediffusion} and its enhanced iteration SDXL~\cite{podell2023sdxl} both employ a pretrained autoencoder for complexity reduction, which comprises an encoder and a decoder. To extend such framework for solving the virtual Try-On problem, it is crucial to maintain consistency in clothing appearance with additional clothing image condition input. Therefore, the input clothing image is fed into the encoder to extract its corresponding features in the latent space. Subsequently, we have engineered a specialized apparel feature processing network, ReferenceNet, which mirrors the architecture of the U-Net~\cite{ronneberger2015unet} found in the original SD model. Both networks were initialized with identical pretrained parameters to ensure consistency. The integration of spatial attention and cross attention layers enabled the successful incorporation of apparel-related features into the denoising pipeline, thereby significantly enhancing the quality of Try-On image generation. 

\begin{figure}[t]
\includegraphics[width=1\linewidth]{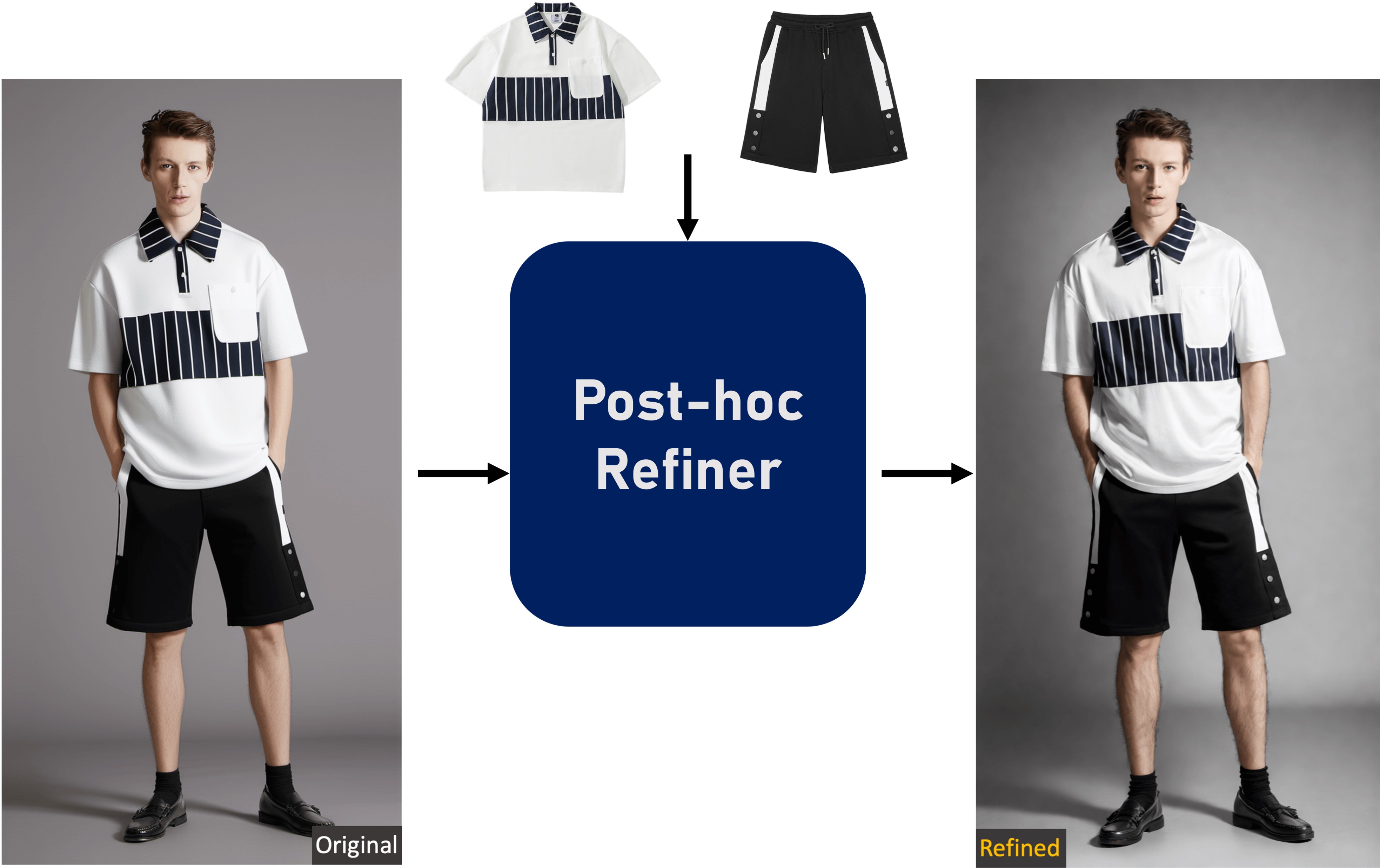}
  \caption{\textbf{Refiner} takes the coarse output from the dual-path conditional diffusion model as its starting point and further enhances it through our subsequent refinement process.
}
\label{fig:inference_pipeline}
\end{figure}

\subsection{Classifier-Free Guidance}
\label{sec:phase31}

In the context of original SD, classifier-free guidance~\cite{ho2022classifier} is a technique used to control the generation process without relying on an external classifier. This method leverages a single diffusion model trained on both conditional and unconditional data. By adjusting the scale of guidance, it can steer the generation process towards producing images that align with a given text prompt. 

In our virtual Try-On framework, we identified the clothing image as the pivotal control element, underscoring its significance over textual prompts. Consequently, we have tailored the unconditional classifier guidance to utilize a blank clothing image, while the conditional guidance is informed by the actual clothing image provided. we are able to harness the guidance scale effectively, thereby delivering more precise and consistent generation outcomes.

\subsection{Background and Lighting Retention}
\label{sec:phase32}
In order to maintain consistency in lighting and background between the generated image and the original image, previous works such as TryonDiffusion~\cite{zhu2023tryondiffusion} have employed a person clothing segmentation model to obtain the clothing mask from the model image. This mask is then slightly expanded, and the corresponding area on the model image is erased. This partially erased image is inputted, and the generation model learns inpainting to fill the clothing area based on this image and the given clothing image. Such approach works well for swapping similar styles of clothing without requiring extensive data. However, it is not suitable for swapping clothes with significant style differences, such as changing from shorts to a long skirt or from tight-fitting to loose clothing. The reason is that the area of the original clothing mask may limit the generation of new clothing and the mask shape might cause undesired coupling with the style.

Our method involves first detecting the bounding box of the person in the model image and then erasing everything except the face and hands. This approach avoids undesired coupling between the mask shape and style, and provides a large enough area to support swapping both upper and lower garments. However, increasing the generated background area might result in significant background differences. In such cases, we can employ a precise person segmentation model to extract the generated person and paste them back into the original background.

\subsection{Pose and Shape Guider}
\label{sec:phase33}
Traditional methods preserved body shape fidelity when swapping single clothing items by selectively replacing image parts, retaining the torso for generating a reasonable body silhouette. However, this approach fails during complete outfit changes, necessitating an additional Pose and Shape Guider for guidance.

In terms of controlling the pose of the person, while previous studies like ControlNet~\cite{controlnet} have demonstrated impressive results, they necessitate additional training phases and parameters. In contrast, we have embraced a more streamlined architectural approach. As detailed in Section \ref{sec:phase32}, to ensure a consistent background and lighting, we incorporated condition images that reflect the pose and shape. All these components can be concatenated together and fed into a denoising U-Net as input. The control images could include skeleton images, dense pose images, or images rendered using the SMPL~\cite{loper2015smpl} model that correspond to the target image. In our experiments, we achieved pose and shape control effects similar to ControlNet, all without the need for additional parameters or training stages.

\begin{figure*}[htpb]
\centering
\includegraphics[width=1.0\linewidth]{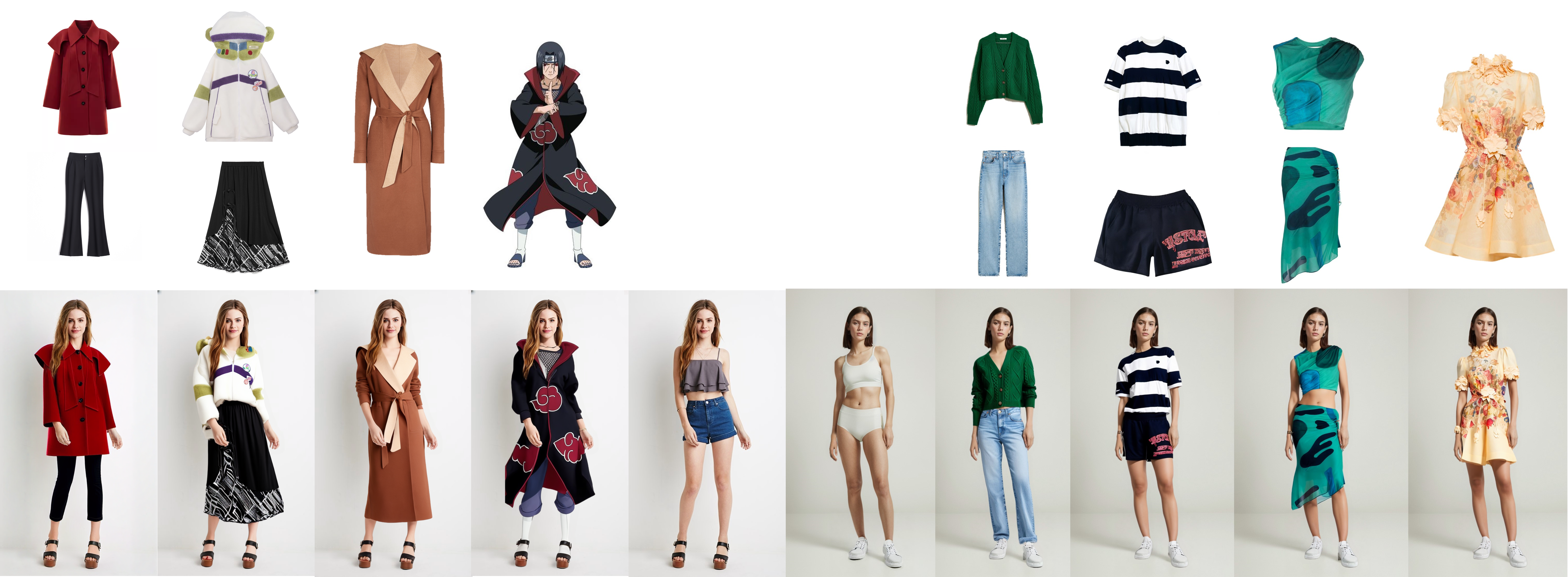}
  \caption{Virtual Try-On with different Outfits.}
\label{fig:diff_outfits}
\end{figure*}

\subsection{Detail Refiner}
\label{sec:phase34}

In our pursuit to create a virtual try-on experience for any clothing and any person, we aimed to incorporate a diverse range of clothing styles and human subjects in our dataset. However, ensuring variety in the dataset while maintaining high image quality and detail proved challenging. To address this, we selected the highest-quality images from the dataset and paired them with model-generated images from the initial version of the virtual try-on, which lacked clear and high-quality details. See \cref{fig:inference_pipeline} for schematic diagram.

By doing so, we constructed a task-specific dataset comprising pairs of high- and low-quality images. Subsequently, we repurposed the virtual try-on framework, employing low-quality images as input and their high-quality counterparts as targets, to train the diffusion model in recovering fine, realistic details effectively.

\section{Results}

In this section, we demonstrate the robust performance of our method, which supports single and multi-piece virtual outfit changes for any clothing, shape, person, and background variations. Remarkably, our technique extends its capabilities to facilitate virtual outfit alterations on animated figures not originally included in our training datasets.

\subsection{Any Outfits}
As shown in \cref{fig:diff_outfits}, OutfitAnyone not only supports single-item clothing virtual try-on, but also allows simultaneous changes for complete outfits, including upper and lower garments. Furthermore, it effectively generates appropriate and realistic try-on results for various clothing styles, including long and short-sleeved tops, trousers and shorts, as well as dresses and similar garments. Compared to prior approaches, OutfitAnyone demonstrates superior adaptability and efficacy in managing an extensive variety of clothing styles and ensembles.

\subsection{Any Person}

\begin{figure}[htpb]
\includegraphics[width=1.0\linewidth]{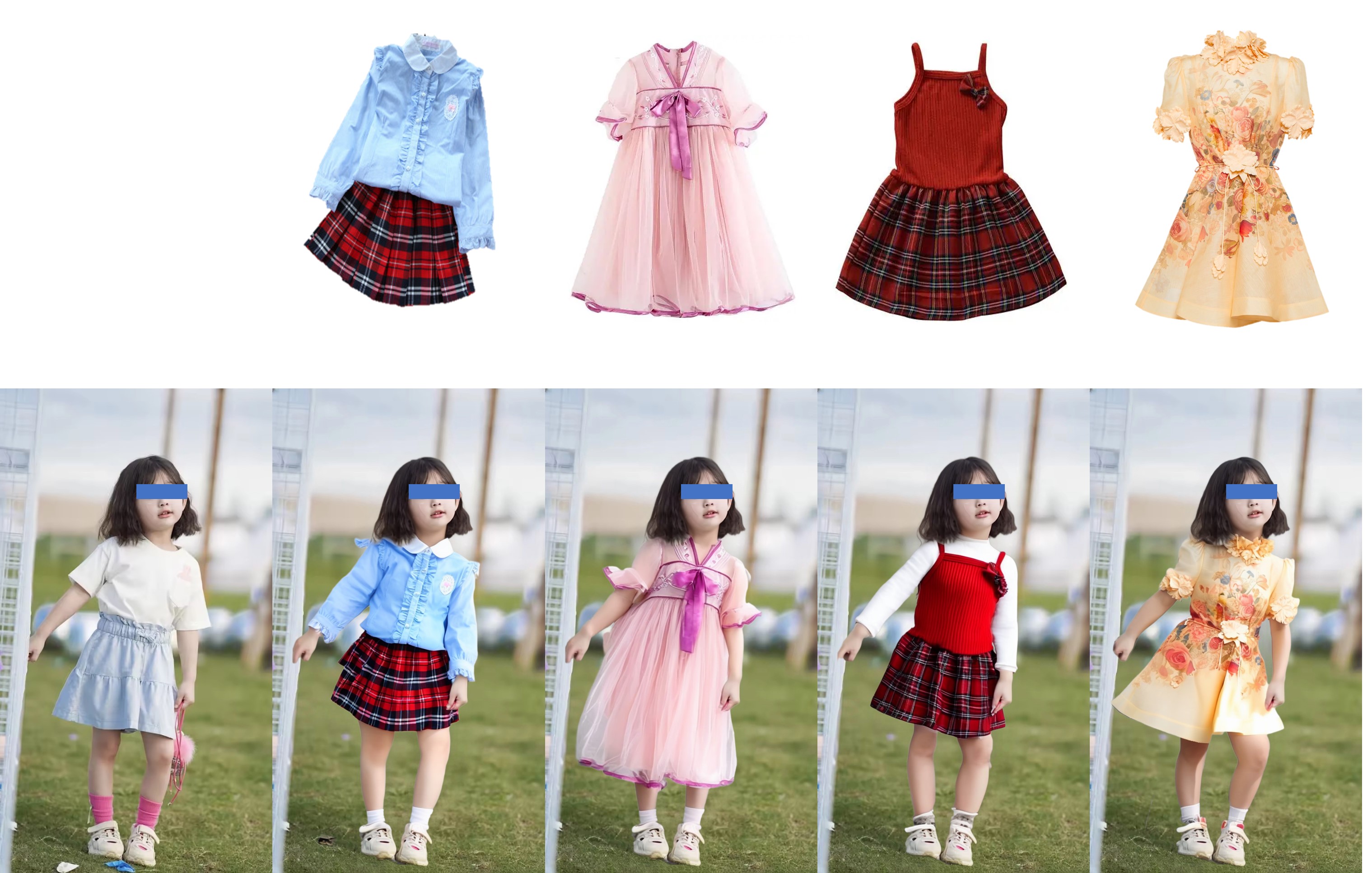}
  \caption{Virtual Try-On for kids.}
\label{fig:kid}
\end{figure}

\begin{figure}[htpb]
\includegraphics[width=1.0\linewidth]{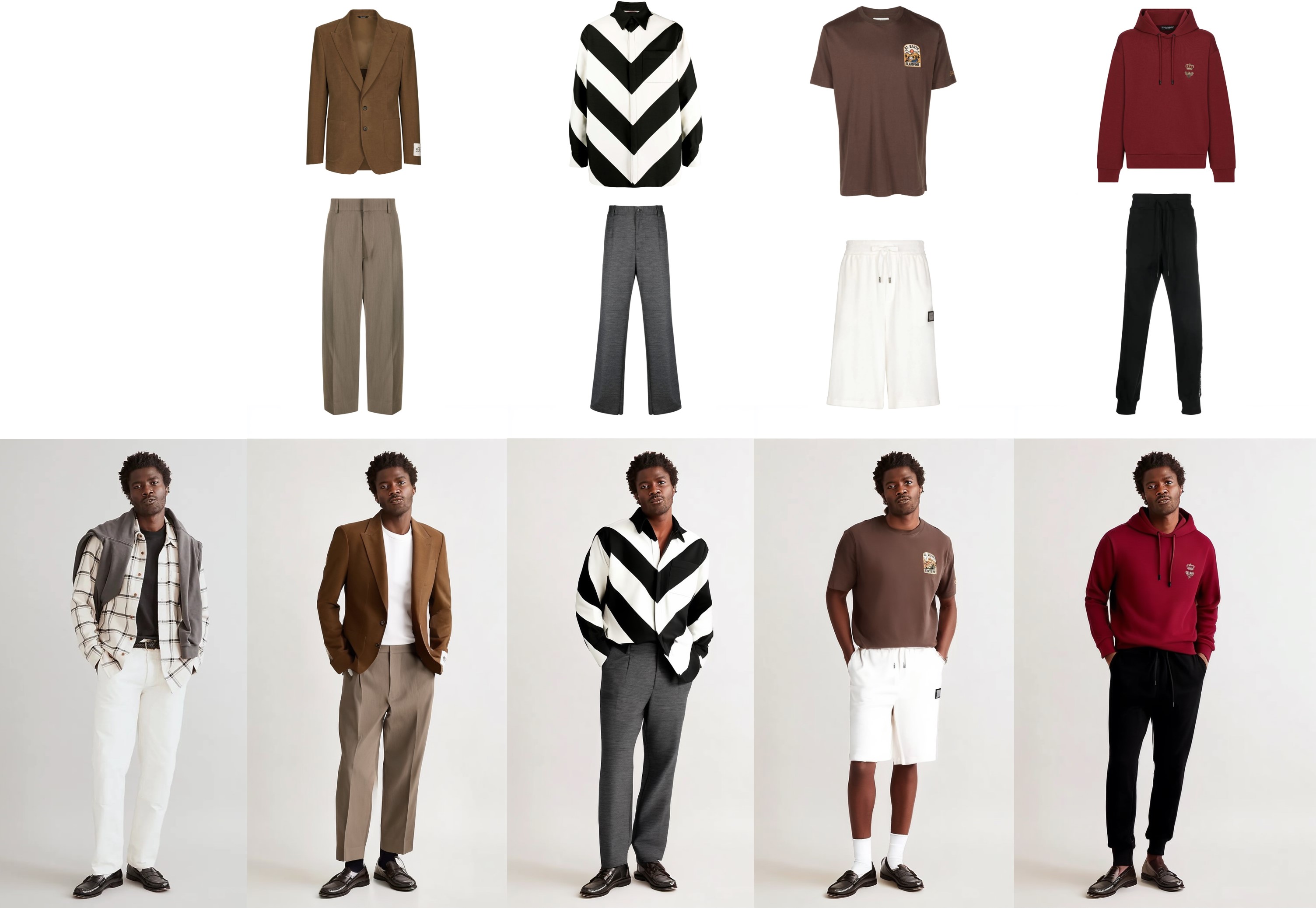}
  \caption{Virtual Try-On for brown-skinned people.}
\label{fig:black_man}
\end{figure}

OutfitAnyone rightly caters to virtual try-on for models of diverse skin tones, ages, and genders, as illustrated in \cref{fig:kid} and \cref{fig:black_man}.  Moreover, it adeptly handles selfie images from everyday users, which often vary greatly in quality and lighting from professional model photos. Despite these differences, OutfitAnyone consistently delivers convincing outfit transformation results, as shown in the final column of \cref{fig:compare}.

\begin{figure}[htpb]
\includegraphics[width=1.0\linewidth]{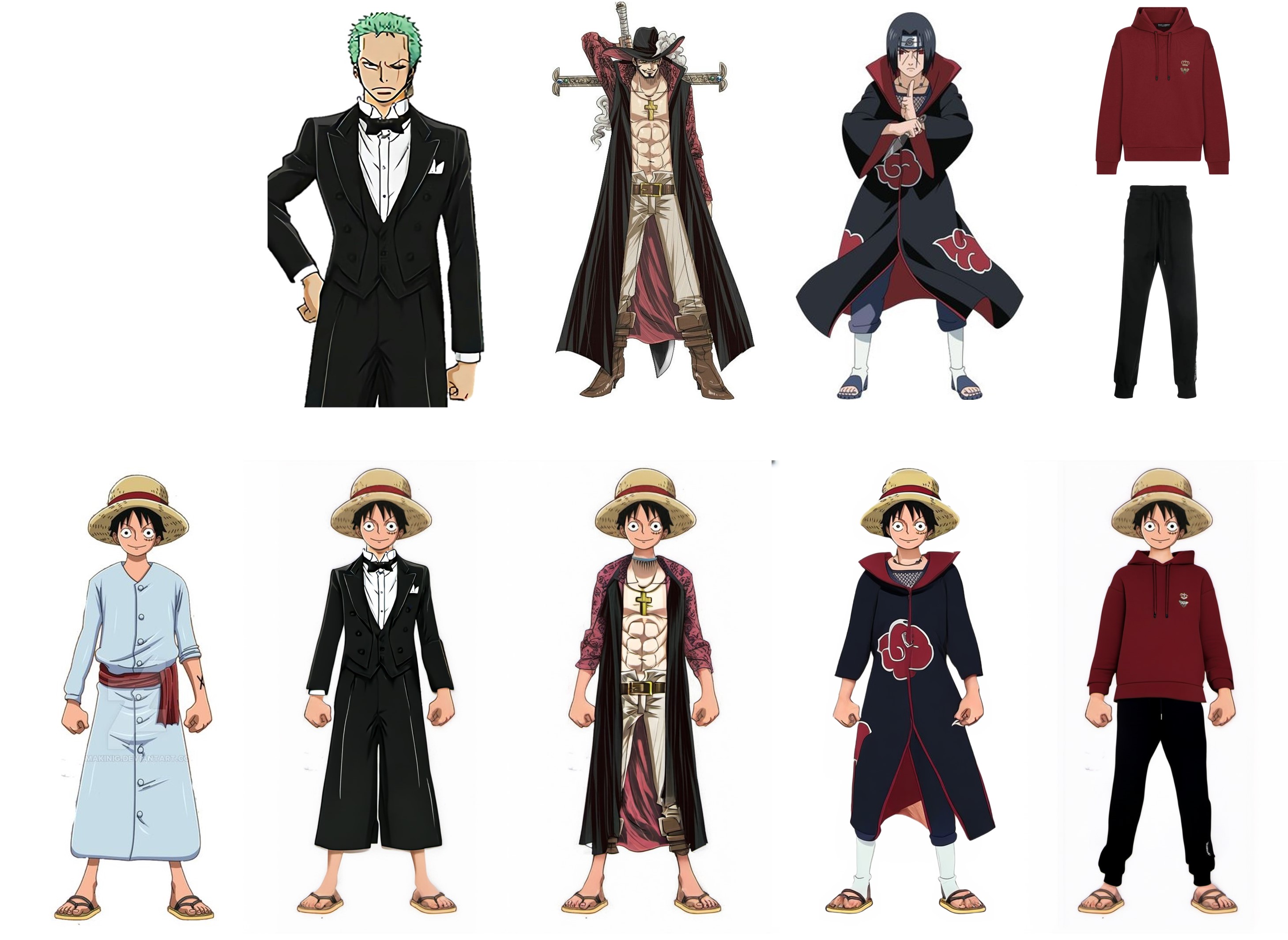}
  \caption{Virtual Try-On for anime character.}
\label{fig:anime}
\end{figure}

Furthermore, our technology extends its prowess to animated characters not included in our training data, as showcased in \cref{fig:anime}. This capability underscores that our model transcends mere rote learning and mimicry; it has acquired genuine understanding and the intelligent capacity to apply outfit changes effectively across various contexts!

\begin{figure}[htpb]
\includegraphics[width=1.0\linewidth]{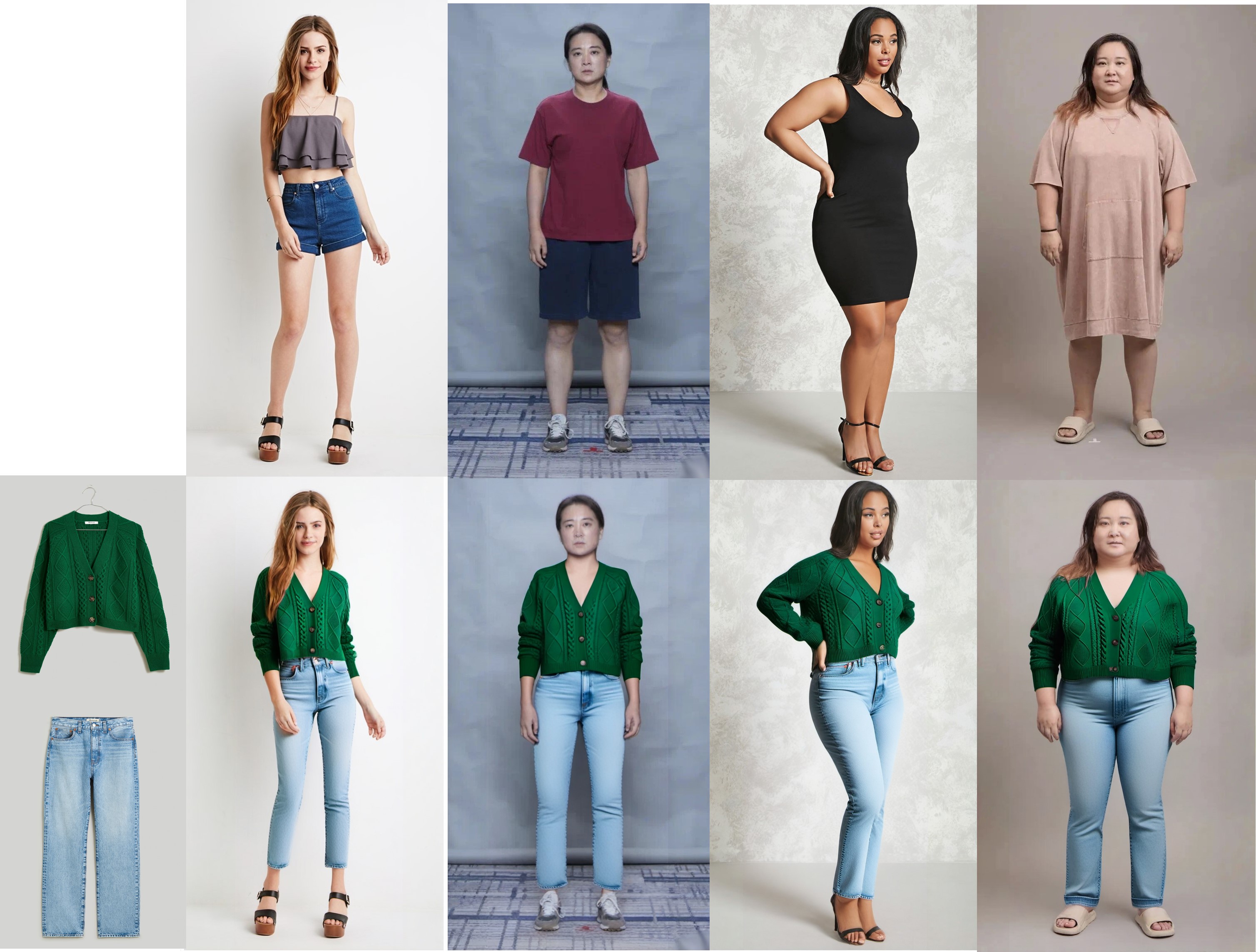}
  \caption{Virtual Try-On for different body shape.}
\label{fig:shape}
\end{figure}

\begin{figure}[htpb]
\centering
\includegraphics[width=0.9\linewidth]{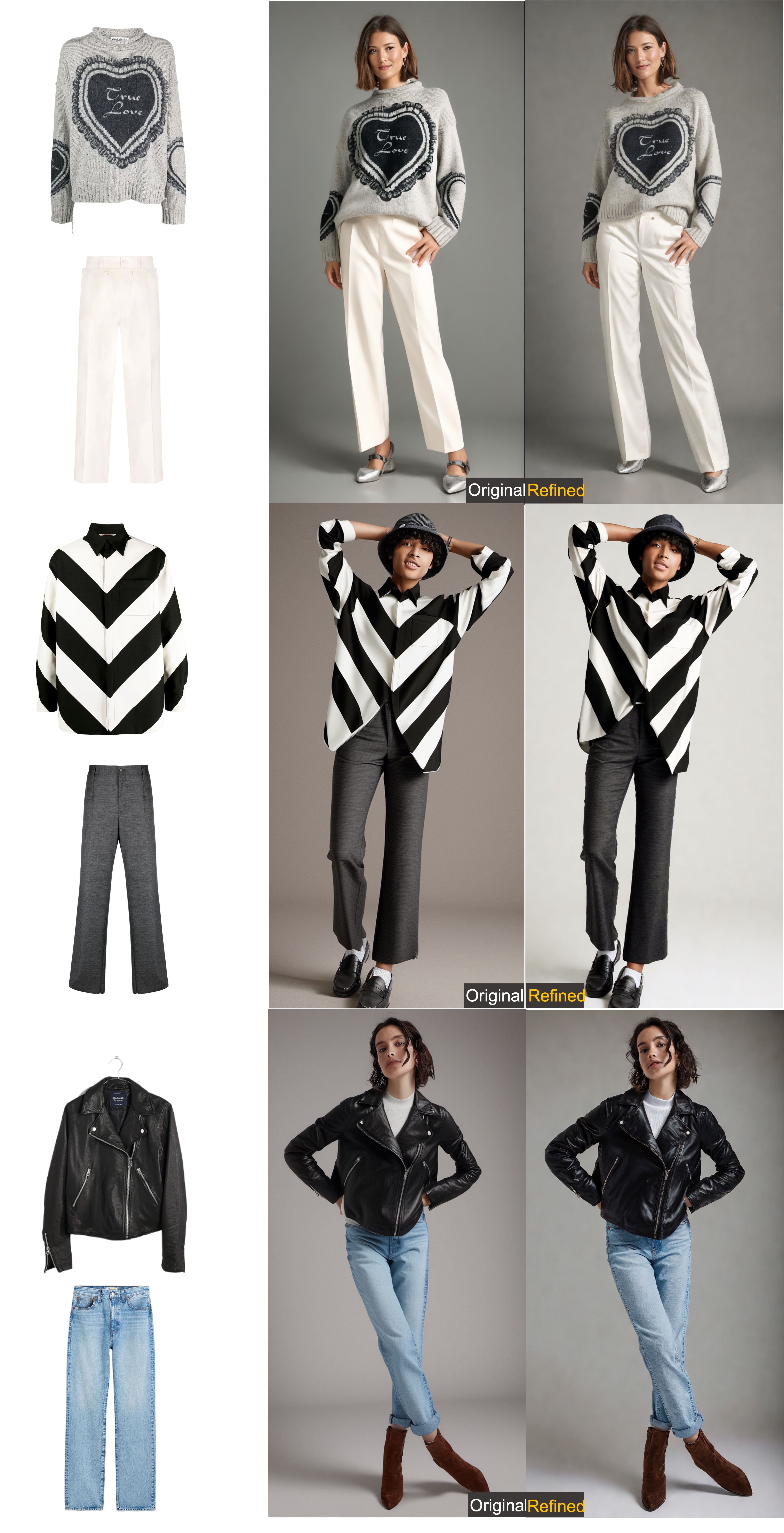}
  \caption{The Refiner model significantly enhances the realism compared to the original model.}
\label{fig:refiner}
\end{figure}

\begin{figure*}[htpb]
\includegraphics[width=1.0\linewidth]{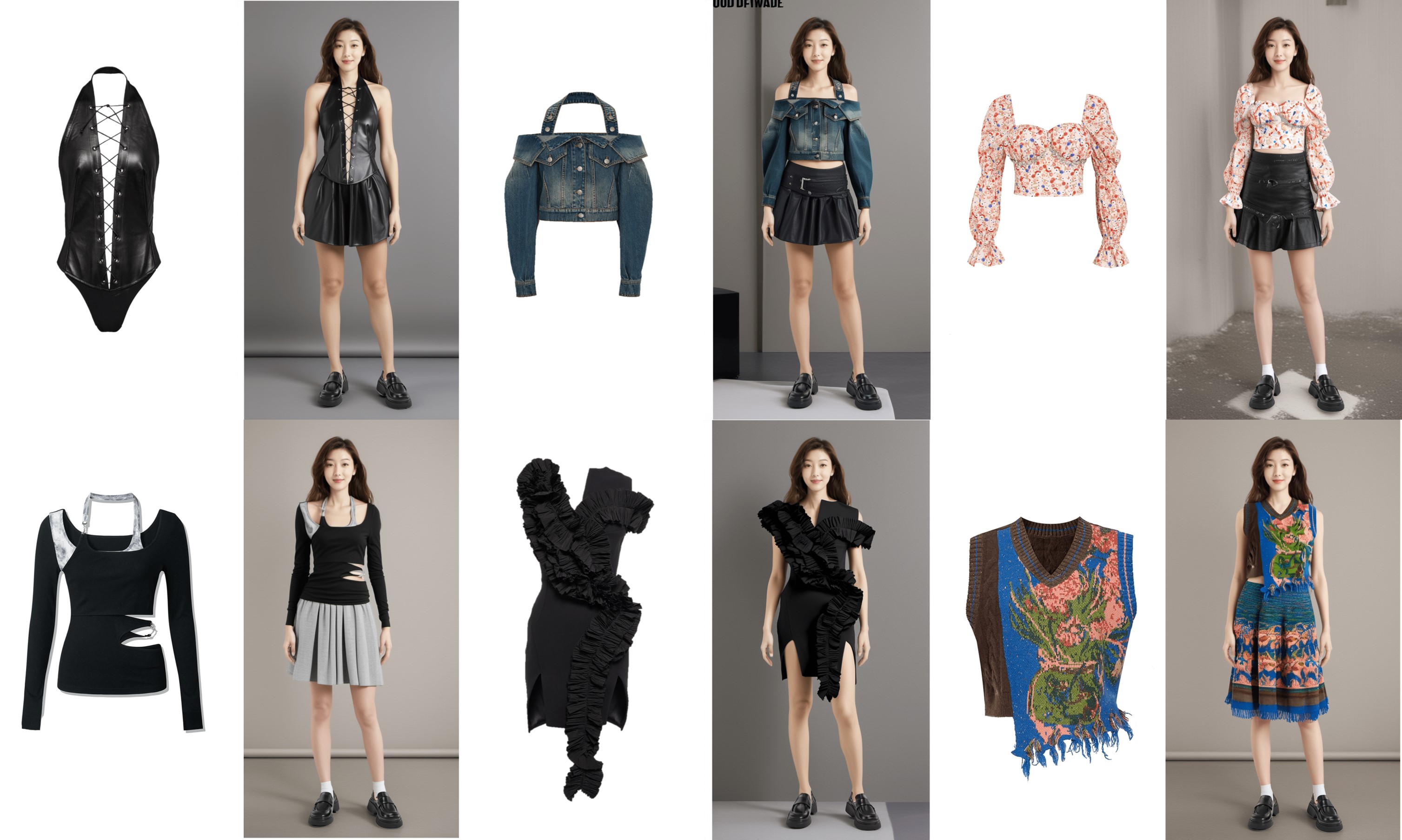}
  \caption{Virtual Try-On for bizzare fashion.}
\label{fig:bizzare}
\end{figure*}

\subsection{Any Body Shape}
% A distinctive aspect of our OutfitAnyone is its capacity to retain the original model's body shape details during a complete outfit change, a capability that was notably absent in previous methodologies. 
% Traditional approaches were only able to maintain body shape fidelity when swapping out individual clothing items, as they selectively replaced parts of the image, preserving the torso to serve as a reference for generating a plausible body silhouette.
Our framework incorporates an additional channel for pose and shape guidance, which extracts densepose-like data (SMPL, openpose are also supported in our work) that mirrors the body's contours. This information is instrumental in directing the final generated model to replicate the exact body shape of the original image. As demonstrated in \cref{fig:shape}, our method excels at preserving the original model's body shape even after a comprehensive outfit change, across a variety of body shapes.

\subsection{Any Background}
\begin{figure*}[htpb]
\includegraphics[width=1.0\linewidth]{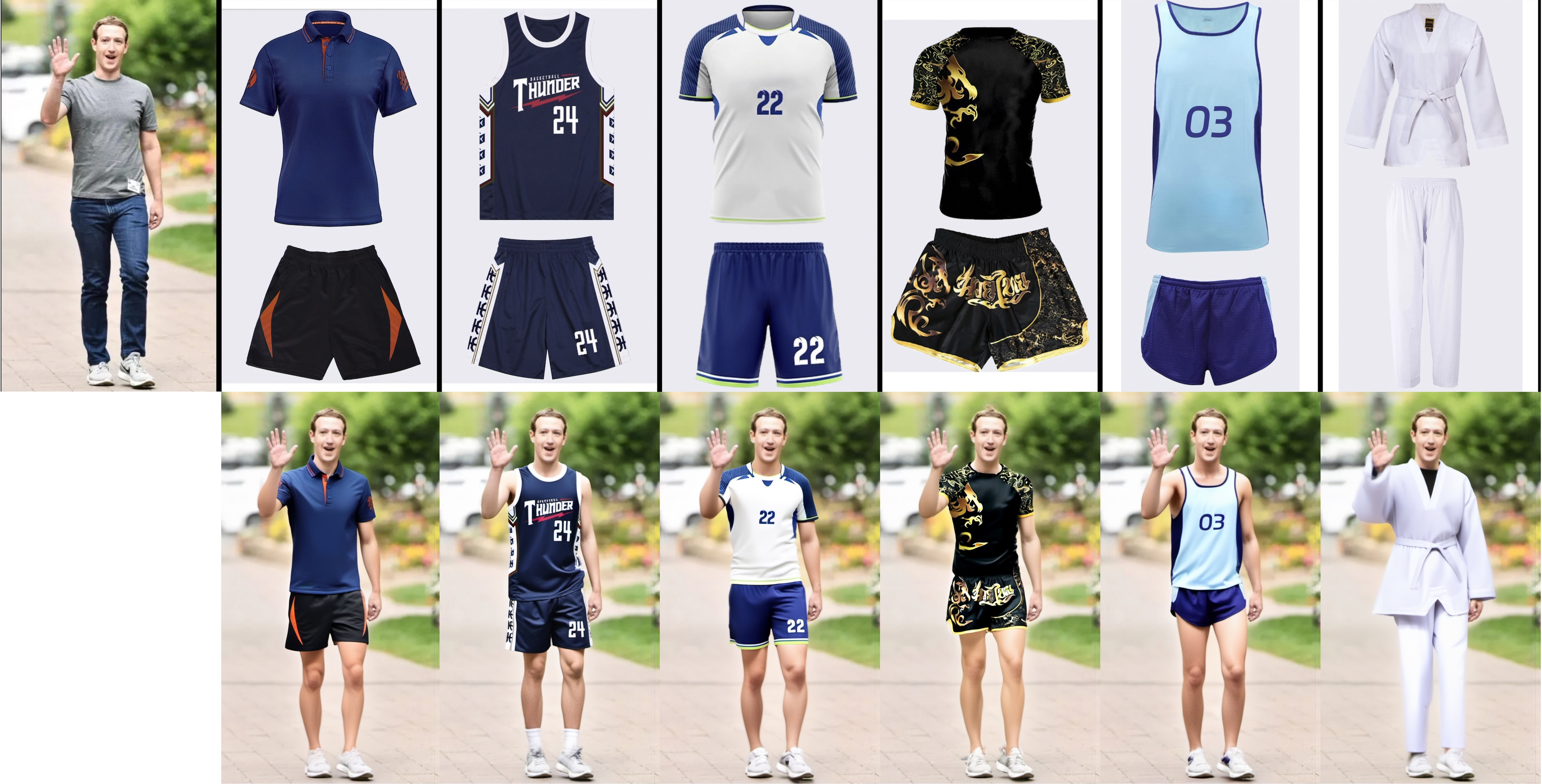}
  \caption{Virtual Try-On for ordinary outdoor image.}
\label{fig:usr2}
\end{figure*}

OutfitAnyone demonstrates exceptional robustness across a variety of backgrounds and lighting scenarios. It generates reasonable clothing lighting effects in complex outdoor scenes, maintaining good performance across diverse indoor and outdoor backdrops, as evidenced in \cref{fig:kid}, \cref{fig:usr2} and the final column of \cref{fig:compare}. This adaptability demonstrates its effectiveness under various environmental conditions and real-world contexts.

\FloatBarrier

\subsection{Refiner}

\begin{figure}[htpb]
\includegraphics[width=1\linewidth]{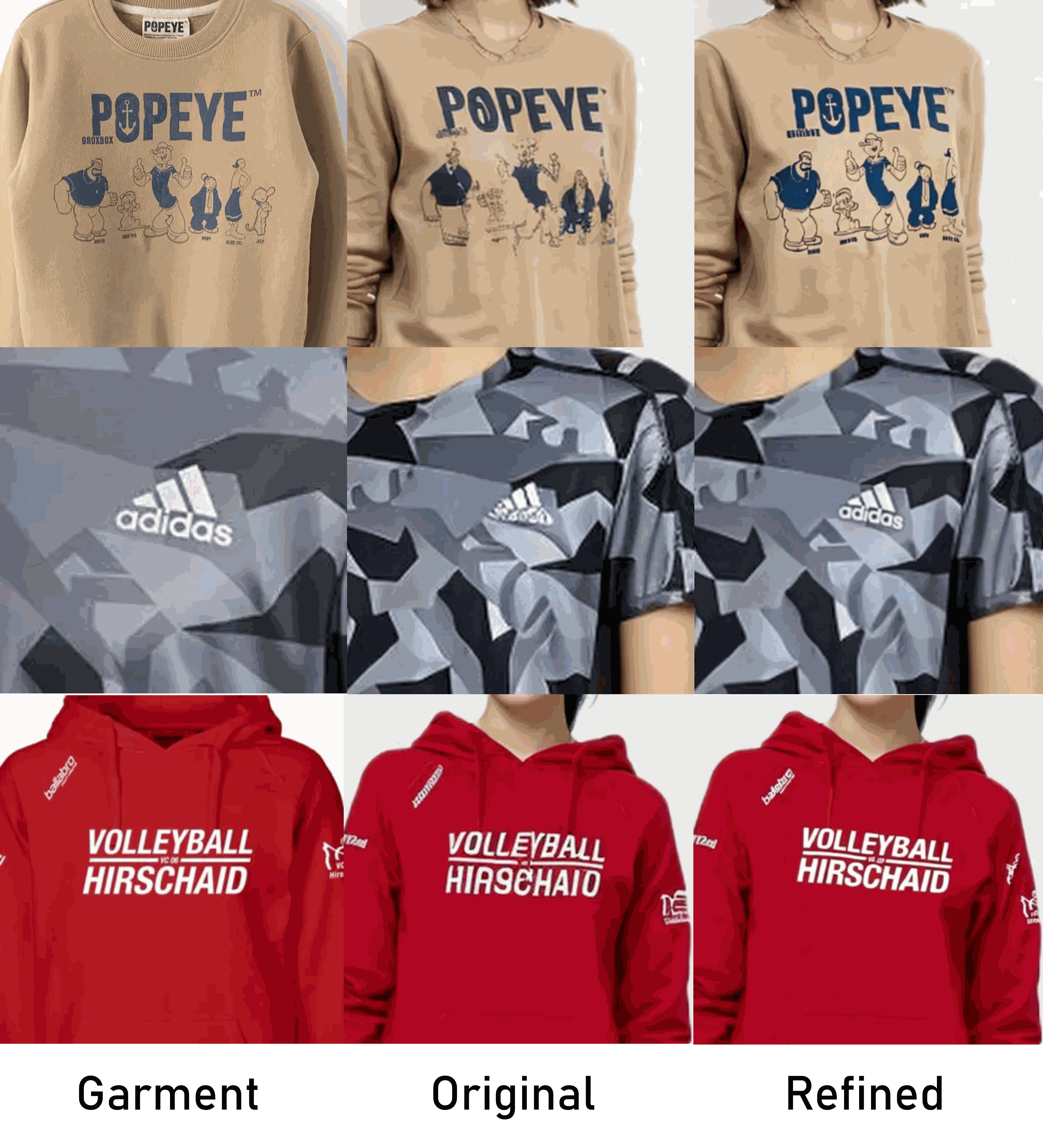}
  \caption{Pre Refiner vs Post Refiner.}
\label{fig:refiner2}
\end{figure}

As mentioned in \cref{sec:phase34}, we proposed utilizing a self-loop refiner model to enhance the realism of virtual try-on results. \cref{fig:refiner} shows that this refiner model significantly boosts the clarity and texture fidelity of the rendered images, while \cref{fig:refiner2} underscores its ability to preserve sharp, localized details. This additional refinement step in OutfitAnyone is crucial for achieving a more vivid and convincing virtual try-on experience.

\subsection{Comparsion}
\begin{figure}[htpb]
\includegraphics[width=1.0\linewidth]{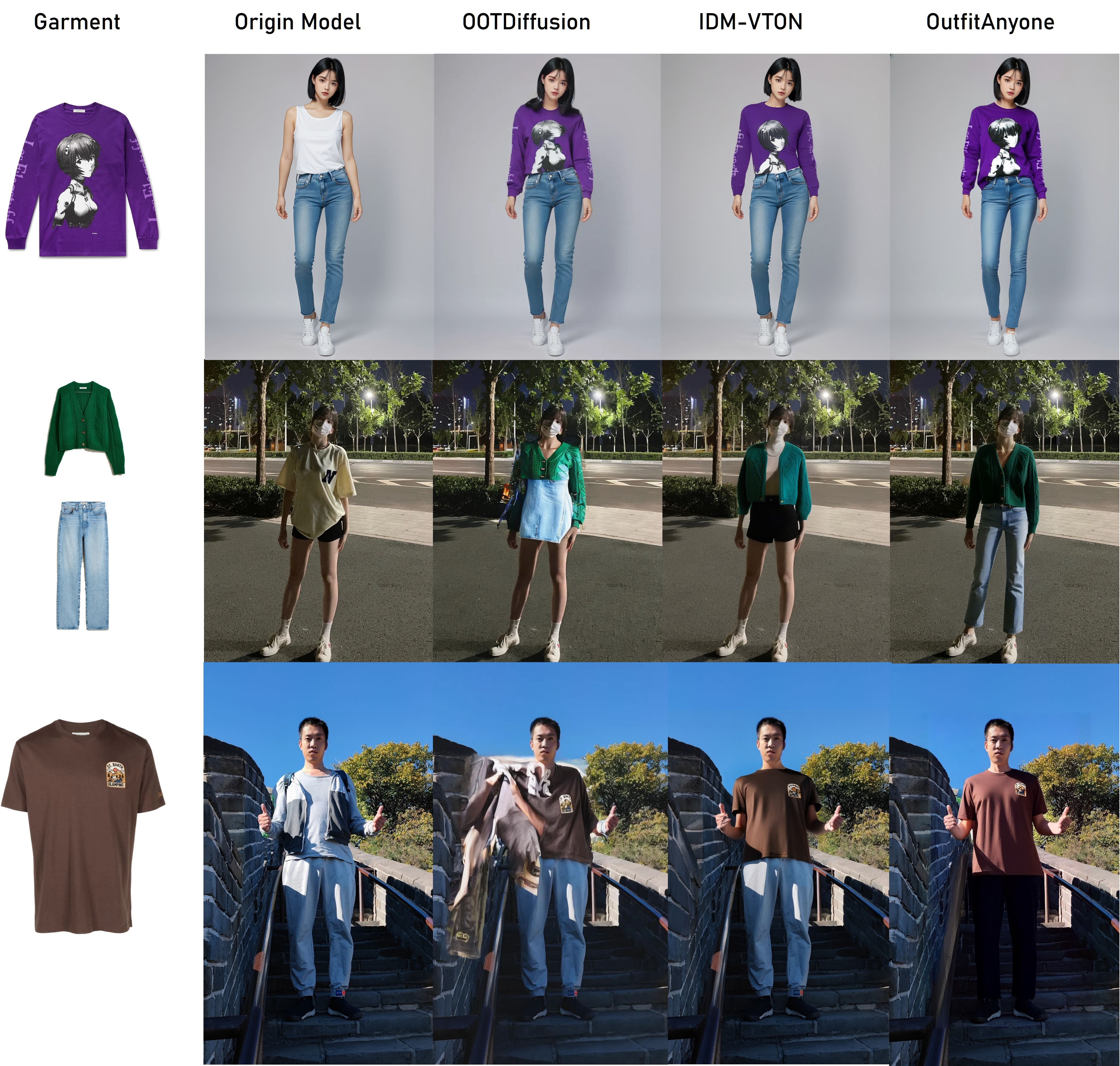}
  \caption{Comparsion betwwen our model and other models in the community. their results come from their huggingface space with the default parameters.}
\label{fig:compare}
\end{figure}

We compared our method with popular community methods like OOTDiffusion~\cite{xu2024ootdiffusion} (with 5k stars on GitHub) and IDM-VTON~\cite{choi2024improving} (with 3k stars on GitHub), our model demonstrated noticeably better performance, particularly in challenging scenarios. As shown in the \cref{fig:compare}, OutfitAnyone excels even when dealing with ordinary users' photos that have complex backgrounds and lighting conditions, which typically make it more difficult to achieve satisfactory virtual try-on results. This highlights the superior robustness of our approach in handling real-world situations, maintaining high-quality performance under various circumstances.

\subsection{Fashion Design Assistor}
Indeed, our OutfitAnyone model proves to be a versatile and helpful resource for fashion designers. By employing its capabilities in generating unique and trendy clothing designs, it can inspire designers to explore new styles and ideas. Moreover, when provided with only a single upper garment, our model can generate suggestions for potential lower garment designs, offering additional creative possibilities and facilitating the design process, As shown in the \cref{fig:bizzare}. Although there are minor details that require attention, we believe that with an increase in training data and optimization of the model, even better results can be achieved.
\section{Conclusion}

Since its initial release at the end of 2023, OutfitAnyone has undergone several iterations, building upon versions \textbf{SD 1.5} and \textbf{SDXL}. Its original open-source version has ranked 14th on the huggingface space, placing it in the top \textbf{0.01}\% of the entire huggingface space (among over 200,000+ projects) \url{https://huggingface.co/spaces/HumanAIGC/OutfitAnyone}, garnering widespread recognition and attention. We are grateful for the emergence of powerful diffusion techniques (SD, SDXL, DDPM/DDIM/DPM, ControlNet and etc) and Google's seminal exploration into virtual try-on: TryonDiffusion, which has allowed us to carve out a distinctive, unique, and mature path for virtual Try-On. In summary, OutfitAnyone has the honour of providing a benchmark application for the practical deployment of AI-generated content (AIGC).

\clearpage
{
    % \small
    \bibliographystyle{ieeenat_fullname}
    % \balance 
    \bibliography{main}
}

\end{document}